\pdfoutput=1

\documentclass[11pt]{article}

\usepackage[]{acl}

\usepackage{times}
\usepackage{latexsym}

\usepackage[T1]{fontenc}

\usepackage[utf8]{inputenc}

\usepackage{microtype}

\usepackage{inconsolata}

\usepackage{graphicx}
\usepackage{multirow}
\usepackage{amsmath}

\newcommand{\ours}[1]{\textsc{LMCor}}
\DeclareMathOperator*{\argmax}{arg\,max}

\usepackage{listings}
\usepackage{xcolor}

\definecolor{Box1Color}{RGB}{227, 236, 246}
\definecolor{Box2Color}{RGB}{248, 220, 225}
\definecolor{Box3Color}{RGB}{255, 238, 224}
\definecolor{Box4Color}{RGB}{232, 228, 240}

\title{Small Language Models Improve Giants by Rewriting Their Outputs}

\author{Giorgos Vernikos$^{1,2}$\thanks{~~Research conducted during an internship at Google.} 
    \quad Arthur Bražinskas$^3$  \quad
  Jakub Adamek$^3$ \\ \textbf{Jonathan Mallinson$^3$ \quad
    Aliaksei Severyn$^3$  \quad  Eric Malmi$^3$}
\\
$^1$EPFL, \ \ $^2$HEIG-VD / HES-SO, \ \ $^3$Google Research \\
\small \tt georgios.vernikos@epfl.ch \\
\small \tt \{abrazinskas, enkait, jonmall, severyn, emalmi\}@google.com \\
}

\begin{document}
\maketitle
\begin{abstract}
Despite the impressive performance of large language models (LLMs), they often lag behind specialized models in various tasks. LLMs only use a fraction of the existing training data for in-context learning, while task-specific models harness the full dataset for fine-tuning.
In this work, we tackle the problem of leveraging training data to improve the performance of LLMs without fine-tuning. Our approach directly targets LLM predictions without requiring access to their weights.
We create a pool of candidates from the LLM through few-shot prompting and we employ a compact model, the LM-corrector (\ours{}), specifically trained to merge these candidates to produce an enhanced output. Our experiments on four natural language generation tasks 
demonstrate that even a small \ours{} model (250M) substantially improves the few-shot performance of LLMs (62B), matching and even outperforming standard fine-tuning. Furthermore, we illustrate the robustness of \ours{} against different prompts, thereby minimizing the need for extensive prompt engineering. Finally, we show that \ours{} can be seamlessly integrated with different LLMs at inference, serving as a plug-and-play module to improve their performance.
\end{abstract}

\begin{figure}[t]
    \centering
    \includegraphics[width=\columnwidth]{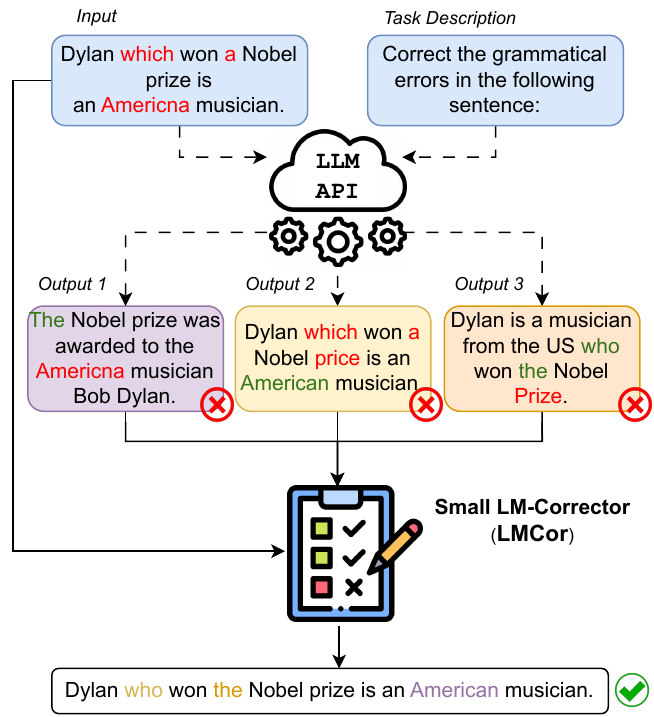}
    \caption{An illustration of our approach for grammatical error correction. We first prompt an LLM to generate multiple outputs via an API (dotted lines). Then we feed the generated candidates to the LM-corrector, a small model that is trained to rewrite them in order to generate the target sentence (solid lines).}
    \label{fig:corrector}
\end{figure}

\section{Introduction}
\label{sec:introduction}

Large language models have recently demonstrated near state-of-the-art performance on various tasks via in-context learning, which enables them to generate outputs based on instructions and a handful of examples, without task-specific training~\citep{brown2020language, GPT3, palm}. However, the effectiveness of this paradigm can vary significantly depending on the task instruction~\citep{shin-etal-2020-autoprompt, jiang-etal-2021-know, schick-schutze-2021-just}, the quantity, relevance and even the order of the in-context examples~\citep{GPT3, gao-etal-2021-making, liu-etal-2022-makes, zhang2023prompting, lu-etal-2022-fantastically}. As a result, in-context learning often requires labour-intensive prompt engineering which does not always guarantee improved performance~\citep{jiang-etal-2021-know}.  

Fine-tuning, on the other hand, has been proven highly effective when task-specific datasets are available, with smaller, fine-tuned models outperforming few-shot-prompted LLMs on various tasks~\citep{lester-etal-2021-power, palm, xu2023small}. While LLMs can also be fine-tuned to enhance their performance in specific tasks, there are several limitations. Firstly, the fine-tuning process can negatively impact the few-shot performance of LLMs on other tasks, leading to a trade-off between versatility and performance~\cite{fu2023specializing}. Secondly, the increasing scale of LLMs makes fine-tuning on standard hardware computationally infeasible. To address these issues, parameter-efficient fine-tuning methods have been proposed~\citep{pmlr-v97-houlsby19a,lester-etal-2021-power,li-liang-2021-prefix, hu2022lora}.
Although these methods are more computationally efficient, they still require access to the model weights and substantial computational resources for loading and updating the model. Furthermore, due to the commercialization of LLMs, they are often available only through restricted inference APIs. 

In light of these challenges, we propose a method that leverages only the outputs of LLMs to enhance their performance. Our work targets scenarios where training data is available, but extreme computational resources are not. To this end, we introduce LM-Corrector (\ours{}), a compact model that corrects the predictions produced by the LLM. Unlike fine-tuning methods, our approach operates directly on the LLM outputs, bypassing the need for access to their weights.

\begin{figure}[t]
    \centering
    \includegraphics[width=\columnwidth]{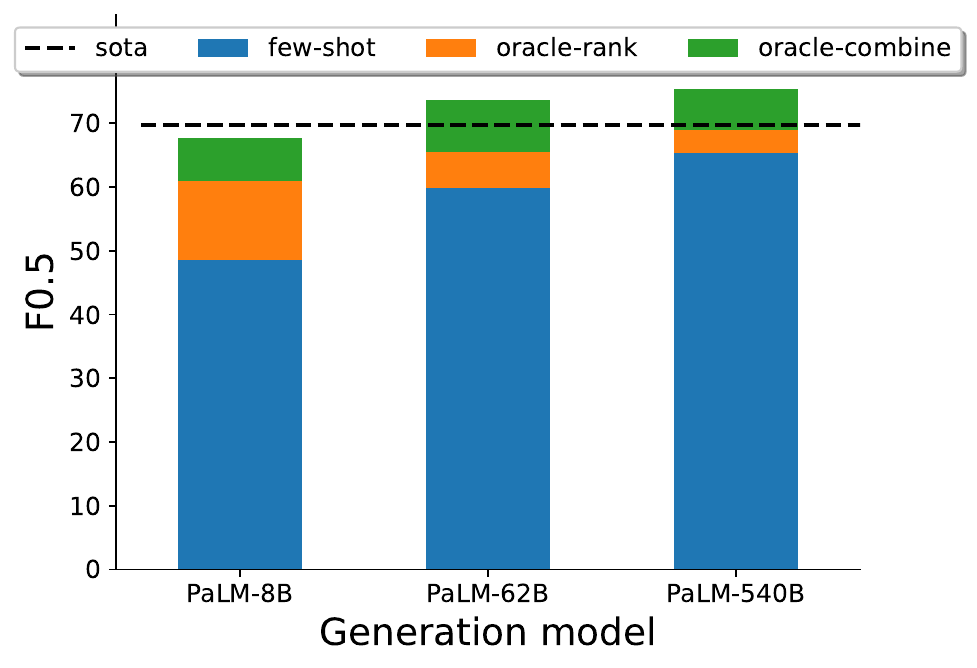}
    \caption{Potential of ranking (oracle-rank) and combining (oracle-combine) sampled candidates (k=10) from PaLM models of different scales for GEC.}
    \label{fig:ranking}
\end{figure}

\ours{} capitalizes on the observation that LLMs can generate a diverse array of candidates for a single input which are often complimentary. Thus, it is possible to produce a superior output by optimally combining spans from different candidates  (see Figure~\ref{fig:ranking}). \ours{} receives multiple candidates for a single input and learns to optimally rank, combine, and edit them, ultimately yielding more precise and higher-quality outputs. Figure~\ref{fig:corrector} illustrates our approach, where \ours{} rewrites the first output of the LLM while incorporating correct spans from the second (\textit{American}) and the third outputs (\textit{the Nobel}) to produce the final, corrected output.

Our contributions can be summarized as follows. (1) We introduce \ours{}, a method to improve the performance of LLMs in the presence of training data without access to the model weights. (2) We conduct experiments on four natural language generation tasks where LLMs underperform specialized models. We demonstrate that a small \ours{} model with only 250 million parameters improves the performance of an LLM with 62 billion parameters, matching or even outperforming task-specific models. (3) We showcase that the corrector is robust to different prompts, alleviating the need for extensive prompt engineering. (4) We demonstrate the versatility of our approach showing that a single corrector can be effortlessly applied to different LMs as a plug-and-play module during inference.
We make our code publicly available\footnote{\url{https://github.com/GeorgeVern/lmcor}}.

\section{Correcting the Outputs of LLMs} \label{sec:method}
In this section we present our computationally efficient approach that utilizes a small model, \ours{}, to correct the predictions of an LLM for a specific task. Unlike traditional fine-tuning methods, our approach does not require access to the weights of the LLM. Instead, as seen in Figure~\ref{fig:corrector}, we interact with the LLM only through an API, as is the case for some state-of-the-art commercial LLMs. 

\paragraph{Headroom analysis}
Our approach is based on the insight that LLMs can generate a diverse pool of candidates for each input, with complementary strengths and weaknesses. Thus, an improved output can be produced by combining the correct parts of the corresponding candidates. To illustrate this, we experiment on the task of grammatical error correction (GEC) ~\cite{ng-etal-2014-conll} using  PaLM models \cite{palm} of varying size, depicted in Figure~\ref{fig:ranking}. 
First, we observe that the few-shot PaLM models underperform fine-tuned 11B-parameter state-of-the-art (sota) GEC model~\cite{rothe-etal-2021-simple}. However, by sampling 10 times from the LLM and employing an oracle to rank the samples (oracle-rank) or to combine correct spans (oracle-combine\footnote{ 
For the oracle-combine we compute the differing spans between the candidates and for each span we choose the one that has the smallest edit distance with the target.}), 
we obtain significant improvements, surpassing state-of-the-art.

This finding highlights the potential of leveraging multiple generations through ranking or combinations to enhance the performance of the LLM via task-specific training.
Motivated by this, we employ a smaller model, \ours{}, to predict the target given the original input and multiple candidates provided by the LLM.

\paragraph{Generating the candidates}

To start our pipeline, we first generate predictions from the LLM via in-context learning. Specifically, we prompt the model with a source sequence $x$, a verbal description of the task $d$, and a handful of demonstrations $e$, depicted as dashed lines in Figure~\ref{fig:corrector}. By sampling from the LLM with a temperature we obtain a diverse set of $k$ candidates, $C=\{c_1, ..., c_k\}$:
\begin{equation}
c_i \sim p_{LLM}(c|x, d, e) \; \; \;  \forall i = 1,2,..,k. \
\end{equation}
\paragraph{Correcting the candidates} 
Next, we feed the set of candidates along with the input sequence $x$ to the corrector\footnote{To indicate end-of-sequence boundaries for the input and the candidates, we use a sentinel token: $x [s] c_1 [s] c_2 [s] ... [s] c_k$.} to generate the final, refined output. 
\begin{equation} \label{eq:corr_input}
\hat{y} = \argmax_{y} p_{\ours{}}(y|x, C)
\end{equation}
In order to train the corrector we fine-tune a small LM on the task-specific dataset augmented with candidates sampled from the LLM. Through this process, \ours{} learns to select the most promising among the generated outputs, combine different candidates and even make necessary edits to compose the desired target sentence.  As we show in the following sections, even a small corrector can substantially improve the quality of LLM outputs, outperform standard fine-tuning, and reduce LLM sensitivity to different prompts.

\section{Experiments \& Results} \label{results}
\subsection{Datasets and Models}
We evaluate \ours{} on four natural language generation tasks: grammatical error correction on CoNLL-14 ~\cite{ng-etal-2014-conll},  data-to-text generation on E2E NLG~\cite{novikova-etal-2017-e2e}, summarization on XSum~\cite{narayan2018don} and machine translation on the English to German translation task from WMT22~\cite{kocmi-etal-2022-findings}.

In most of our experiments, we use the 62B version of PaLM~\cite{palm} as our large LM except for Section~\ref{sec:size} where we vary the size of the LLM up to 540B parameters. For the machine translation task we use the 2.9B version of XGLM~\cite{lin-etal-2022-shot} as our LLM, since at the time of running this experiment, it was more easily accessible to the first author. We prompt the LLM with a task description and a number of demonstrations randomly selected from the respective validation set. We sample \textit{k} = 4 times from the LLM with a temperature of 0.7 for PaLM and employ nucleus sampling~\cite{Holtzman2020The} with $p=0.6$ and a temperature of 0.6 for XGLM. Additionally, we include the greedy-decoded output as a candidate since initial results showed that it improves performance.  We use T5.1.1\footnote{\href{https://github.com/google-research/text-to-text-transfer-transformer/blob/main/released_checkpoints.md}{https://github.com/google-research/text-to-text-transfer-transformer/blob/main/released\_checkpoints.md}} base~\cite{t5}  (250M parameters) as our model both for the \ours{} and the standard fine-tuning baseline. 
We choose the model based on the performance on the validation set. The outputs of the corrector and the T5 baseline are generated via beam search with a beam of size 5.

We compare our approach, \ours{}, with the following baselines: 1) in-context learning using the LLM (ICL), prompted with the same number of demonstrations, and 2) standard fine-tuning with a T5-base and PaLM. We also provide scores for 3) the reranking approach of \citet{mbrd} where they use Minimum Bayes Risk Decoding (MBRD) combined with an alignment function combined to select one among the candidates produced from the LLM. We use the same pool of candidates that are used as input to the corrector and employ Sim-LCS, a lexical similarity function  based on longest common subsequence which achieved the best results across tasks among the alignment functions. We additionally provide the scores of an oracle reranker that selects the candidate with the smallest edit distance compared to the target as an upper-bound of reranking methods. Finally, we provide the results for a version of our approach that feeds the corrector with only the greedy-decoded candidate (single).

\subsection{Grammatical Error Correction}\label{sec:exp_gec}

Grammatical Error Correction (GEC) is a text-to-text task that requires correcting the grammatical errors while applying minimal changes to the original input sentence. Despite being trained on vast amounts of text, LLMs have been demonstrated to underperform task-specific models in this task~\cite{yasunaga-etal-2021-lm, suzgun-etal-2022-prompt}.  

We use the CoNLL-14~\cite{ng-etal-2014-conll} dataset as our testset. Following previous work~\cite{rothe-etal-2021-simple}, we use the combination of the FCE~\cite{yannakoudakis-etal-2011-new} and W\&I~\cite{bryant-etal-2019-bea} datasets (60k examples) for training and validation. We report $F_{0.5}$ scores obtained with the MaxMatch scorer~\cite{dahlmeier-ng-2012-better}\footnote{\href{https://www.comp.nus.edu.sg/~nlp/conll14st.html}{https://www.comp.nus.edu.sg/~nlp/conll14st.html}}. We use 5 demonstrations in the LLM prompt in order to generate the candidates during training and inference.

\begin{table}[t]
\centering
\resizebox{0.55\columnwidth}{!}{%
\begin{tabular}{lc}
\hline
\textbf{Model} & \textbf{F0.5}  \\
\hline
T5-base & 59.38 \\ 
\hline
PaLM-62B (ICL) & 59.92 \\
+ MBRD-Sim-LCS & 58.87 \\
+ Oracle Reranker & 63.88 \\ \hline
+ \ours{} (single)   & \underline{62.47}  \\
+ \ours{} (mult.) & \bf{62.48}  \\ 
\hline
\end{tabular}
}
\caption{Results of our approach in GEC (CoNLL-14). The first group indicates fine-tuned models, the second group in-context learning with reranking and the final group provides the scores for \ours{}. The best scores are in \textbf{bold} and the second best ones are \underline{underlined}.}
\label{tab:conll_results}
\end{table}
\begin{figure}[t]
    \centering
    \includegraphics[width=0.8\columnwidth]{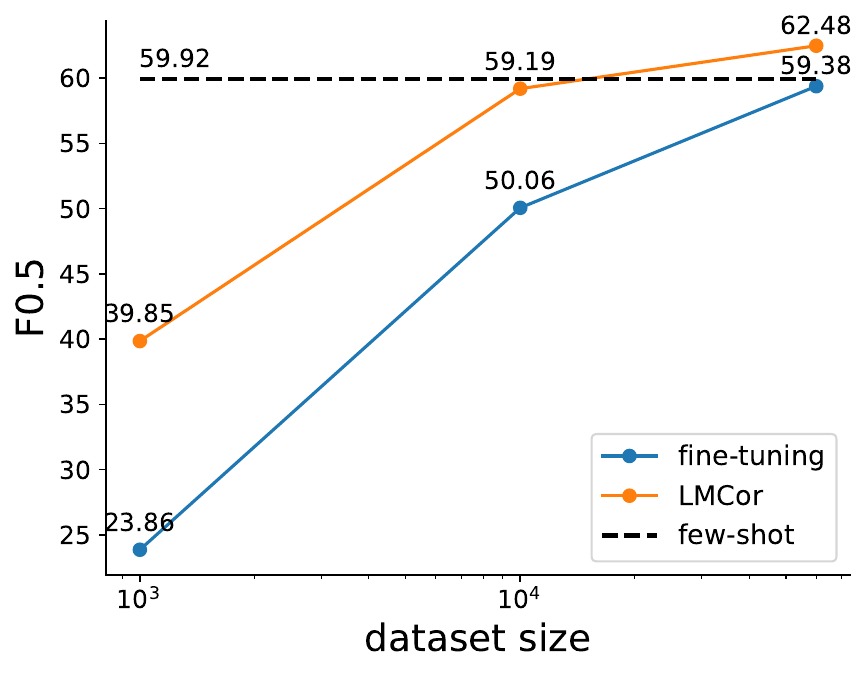}
    \caption{The effect of dataset size for standard fine-tuning and \ours{}. Results are reported on GEC.}
    \label{fig:scaling_data}
\end{figure}
The results presented in Table~\ref{tab:conll_results} show that standard fine-tuning and in-context learning exhibit comparable performance in GEC while our approach significantly outperforms both, by 3 and 2.5 $F_{0.5}$ points respectively. It is worth noting that MBRD does not yield any improvement over standard few-shot prompting. However, as expected the use of an oracle to rank the produced hypotheses results in a considerable performance boost. Although \ours{} does not surpass the performance of the oracle it manages to significantly close the gap, demonstrating the ability of the corrector to identify high-quality candidates from the model.

Additionally, we compare the performance of standard fine-tuning and \ours{} across varying numbers of training instances. As illustrated in Figure~\ref{fig:scaling_data} our approach consistently outperforms the baseline for all dataset sizes. The gap is particularly pronounced when the training dataset is limited, consisting of only 1k examples, resulting in a substantial difference of 15 points in $F0.5$. The sample efficiency of \ours{} can be attributed to its ability to leverage the candidates generated by the LLM to produce more accurate outputs. We note that in this low-resource scenario, both trained models perform worse than few-shot prompting. This outcome is expected as the extensive pretraining of LLMs on language generation enables them to perform grammatical error correction out of the box. As the dataset increases to 10k examples we observe that \ours{} performs on par with the LLM while the baseline continues to underperform. Beyond this threshold, \ours{} surpasses both in-context learning and fine-tuning by utilizing both the training data and the candidates.

\begin{table}[t]
\centering
\resizebox{0.65\columnwidth}{!}{%
\begin{tabular}{lcc}
\hline
\textbf{Model} & \textbf{R-2} & \textbf{R-L} \\
\hline
T5-base & 45.3 & 52.8 \\
PaLM-62B* (FT)  & 45.2 & --\\
PaLM-540B* (FT) & \underline{45.3} & 52.3 \\
\hline
PaLM-62B (ICL) & 35.1 & 45.6 \\
+ MBRD-Sim-LCS & 35.7 & 46.2 \\
+ Oracle Reranker & 37.1 & 50.4 \\ \hline
+ \ours{} (single)   & 44.8 & \underline{52.8}  \\
+ \ours{} (mult.)  & \textbf{45.6} & \textbf{53.4}  \\ 
\hline
\end{tabular}
}
\caption{Results of our approach in E2E NLG (cleaned). Results with \textit{*} are reported from the original paper~\cite{palm}. The first group indicates fine-tuned models, the second group in-context learning with reranking and the final group provides the scores for \ours{}. The best scores are in \textbf{bold} and the second best ones are \underline{underlined}.}
\label{tab:e2e_results}
\end{table}

\subsection{Data-to-text}

The next task we evaluate on is E2E NLG~\cite{novikova-etal-2017-e2e}, a data-to-text task where the input is a number of key-value pairs about a restaurant and the output is a short description of the restaurant in natural language. We use the cleaned version of the dataset, E2E NLG (cleaned)~\cite{dusek-etal-2019-semantic} and the default splits for training (35k examples), validation and testing. We use 5 demonstrations to produce the candidates for the corrector both during training and inference. We report ROUGE-2 and ROUGE-L~\cite{lin-2004-rouge} scores.

Table~\ref{tab:e2e_results} presents a comparison between standard fine-tuning with a T5-base model, in-context learning, and fine-tuning using PaLM models. Notably, standard fine-tuning with a T5-base significantly outperforms in-context learning and achieves results comparable to fine-tuning with the much larger PaLM models\footnote{\citet{palm} attribute the mediocre performance of the fine-tuned PaLM models to the small dataset size and the `\textit{significant mismatch with the pre-training corpus}'.}.
While reranking techniques improve the performance of few-shot prompting, even the oracle approach falls short of the performance achieved by the fine-tuned T5-base model. This highlights the primary limitation of reranking approaches, particularly for challenging tasks for LLMs, where their upper bound solely depends on the quality of candidates. 

In contrast, the performance of \ours{} does not exclusively rely on the quality of the candidates. The corrector module has the ability to edit the LLM-generated candidates, leading to more accurate outputs. As a result, \ours{} demonstrates the best overall performance for the E2E task, surpassing even the fine-tuned PaLM-540B model by 1 point in ROUGE-L. An important characteristic of our approach is the ability of the corrector to observe multiple candidates for a single input. This enables \ours{} to combine candidates in order to compose a more refined answer. This is supported by the performance discrepancy between \ours{} (single) and \ours{} (mult.) highlighting the effectiveness of leveraging multiple candidates.   

\subsection{Summarization}

The third task that we consider is abstractive summarization. Specifically, we use XSum~\cite{narayan2018don} with the default train (204k examples), validation and test splits. Due to the length of the articles we truncate the inputs and only use 1 demonstration to prompt the LLM. To handle the increased sequence length in the input of the corrector we again truncate the articles and use a maximum sequence length of 2048 tokens. We report ROUGE-1, ROUGE-2 and ROUGE-L scores. 

The results of Table~\ref{tab:xsum_results} reveal that standard fine-tuning outperforms in-context learning for the XSum dataset. Specifically,  the fine-tuned T5 and PaLM-62B models outperform in-context learning by 15 and 18 points in ROUGE-2 respectively. The difficulty of the task, which involves summarizing an article into a single sentence, poses challenges for in-context learning, while the substantial dataset size of 204k examples favors fine-tuning. However, the use of a corrector module leads to singificant improvements over in-context learning with the 62B PaLM model, resulting in performance gains of 6 points in ROUGE-2 and 8 points in ROUGE-L. Notably, \ours{} outperforms few-shot learning with the largest 540B PaLM model. Again, the use of multiple candidates by the corrector further enhances the performance of \ours{} confirming our hypothesis regarding the complementarity of the outputs generated by the LLM.

On the other hand, the \ours{} performs slightly worse than standard fine-tuning. This result is somewhat unexpected since the input for the corrector is a strict superset of the model input in the fine-tuning setting. We attribute the slight drop in performance to the poor quality of the provided candidates, which introduces undesirable noise to the corrector's input, a hypothesis that we test in Section~\ref{sec:pegasus}. 
Additionally, the long-range dependencies that are introduced by simultaneously processing the article and the generated summaries might also contribute to the performance gap between \ours{} and standard fine-tuning.

\begin{table}[t]
\centering
\resizebox{0.8\columnwidth}{!}{%
\begin{tabular}{l c c c}
\hline
\textbf{Model}  &\textbf{R-1} & \textbf{R-2} &\textbf{R-L}\\
\hline

T5-base & \textbf{38.64} & 16.98 & 31.41\\
PaLM-62B* (FT) &-- &18.5 &-- \\
PaLM-540B* (FT) &-- &\textbf{21.2} &\textbf{36.5} \\
\hline 
PaLM-62B (ICL) & 28.18 & 10.50 & 22.38 \\
PaLM-540B (ICL) & 29.88 & 11.75 & 23.83 \\
\hline
+ \ours{} (single) & 36.98 & 16.41 & 30.20 \\
+ \ours{} (mult.) & \underline{37.62} & \underline{16.50} & \underline{30.67} \\ 
\hline
\end{tabular}
}
\caption{Results of our approach on XSum. Results with \textit{*} are reported from the original paper~\cite{palm}. The first group indicates fine-tuned models, the second group in-context learning and the final group provides the scores for \ours{}. The best scores are in \textbf{bold} and the second best ones are \underline{underlined}.}
\label{tab:xsum_results}
\end{table}

\subsection{Machine Translation}

The final task in our evaluation is machine translation (MT). For this task we use the English to German language pair from WMT22~\cite{kocmi-etal-2022-findings} as our test set and the corresponding pair from WMT21~\cite{akhbardeh-etal-2021-findings} as our validation set. Our training data consists of 200k examples sampled from the News Commentary v16 corpus\footnote{\url{https://www.statmt.org/wmt22/translation-task.html}}. During both training and inference, we prompt the LLM with 5 demonstrations to generate the candidates. We report scores using traditional surface-based MT evaluation metrics like BLEU~\cite{papineni-etal-2002-bleu}, as well as more recent neural-based metrics such as COMET-22~\cite{rei-etal-2022-comet} and BLEURT~\cite{sellam-etal-2020-bleurt}.

The findings presented in Table~\ref{tab:mt_results} indicate that, similar to previous tasks, standard fine-tuning outperforms in-context learning for MT across two of the three considered metrics. While the minimum Bayes risk reranking approach shows some improvements, it fails to significantly narrow the gap with the fine-tuned baseline. Notably, the oracle reranker manages to surpass T5 in terms of COMET scores, although T5 still outperforms in terms of BLEU.

Our proposed approach achieves substantial gains, surpassing both fine-tuning and in-context learning, as well as the reranking approaches, including the oracle. This observation once again highlights the limitations of reranking approaches, especially when dealing with low-quality candidates. Specifically, the version of \ours{} using a single candidate demonstrates improvements of 4.5 BLEU points, 2 points in terms of COMET, and 0.5 points in terms of BLEURT compared to the best scores achieved either by fine-tuning or reranking. The inclusion of multiple candidates yields further gains ranging from 0.6 to 1.2 points, depending on the metric,  underscoring the benefits of a diverse candidate pool.

\begin{table}[t]
\centering
\small
\begin{tabular}{lccc}
\hline
\textbf{Model} & \textbf{BLEU} & \textbf{COMET} & \textbf{BLEURT}\\ \hline
T5-base & 23.32  &  75.22 & 64.57\\ \hline
XGLM-2.9B (ICL) & 17.32 & 74.54 & 66.47\\ 
+ MBRD-Sim-CLS & 18.01 & 74.82 & 66.73 \\ 
+ Oracle Reranker & 21.21 & 75.55 & 66.90\\ \hline
+ \ours{} (single) & \underline{24.51} & \underline{76.81} & \underline{67.23} \\ 
+ \ours{} (mult.) & \textbf{25.15} & \textbf{77.45} & \textbf{68.41} \\ \hline
\end{tabular}
\caption{Results of our approach on WMT22 En->De. The first group indicates fine-tuned models, the second group in-context learning and the final group provides the scores for \ours{}. The best scores are in \textbf{bold} and the second best ones are \underline{underlined}.}
\label{tab:mt_results}
\end{table}

\section{Robustness Analysis}
\subsection{Different prompts}

We have demonstrated that \ours{} enhances the performance of LLMs by refining their generated predictions. However, the few-shot paradigm remains appealing since it does not require training or access to a dataset, except for a handful of examples used in the prompt. While prompt construction is not computationally intensive, there is no consensus on the optimal selection, number and even order of examples in the prompt which can lead to variations in predictions and significantly impact LLM performance~\cite{lu-etal-2022-fantastically}. To investigate whether \ours{} is similarly affected by this variability we use three different sets of 5 demonstrations to prompt the LLM for the task of GEC. We then feed the generated predictions as input to the corrector. It is important to note that we trained \ours{} only once using the candidates generated with the original set of demonstrations (set 1) and simply swapped candidates during inference.

\begin{table}[t]
\centering
\resizebox{0.9\columnwidth}{!}{%
\begin{tabular}{l|ccc|cc}
\hline
\textbf{Model} & \textbf{set 1} & \textbf{set 2} & \textbf{set 3} & \textbf{mean} & \textbf{std} \\
\hline
PaLM-62B & 59.9 & 58.9 & 56.2 & 58.6 & 1.9  \\ \hline
+ \ours & 62.5 & 62.3 & 62.9 & \textbf{62.6} & \textbf{0.3} \\\hline
\end{tabular}
}
\caption{Mean $F_{0.5}$ score and standard deviation (std) using different sets of demonstrations for in-context learning vs. our approach for GEC.}
\label{tab:variance}
\end{table}

Table~\ref{tab:variance} highlights the significant variance in LLM performance depending on the selection of demonstrations with a difference of $3.7$ between the highest and the lowest $F_{0.5}$ score. In contrast, \ours{} remains unaffected and achieves competitive performance even when the quality of candidates significantly deteriorates (set 3). This demonstrates the capability of \ours{} to compensate for candidates of poor quality by performing edits on them. The robustness of the corrector against prompts of varying quality (with a variance of $0.3$ compared to $1.9$ for the LLM) suggests that it can mitigate the need for extensive prompt engineering.

\subsection{Different LLMs} \label{sec:size}
In the previous experiment we demonstrated the robustness of the corrector against candidates of varying quality. In this set of experiments we further examine the robustness of our approach by testing whether a corrector can be used interchangeably with different LMs without retraining. It is important to note that we only trained the corrector once and performed inference by swapping the LLM responsible for generating the candidates. 

Initially, we focus on LMs from the same family of models, namely PaLM~\cite{palm}, which share similar architectures and training data but differ in the number of parameters. Table~\ref{tab:different_scales} presents the results of applying the corrector to different PaLM models than the one it was originally trained on (62B). Across all scales, the \ours{} consistently outperforms standard fine-tuning and in-context learning, with the exception of the 540B model where the performance is comparable. The corrector achieves significant gains when applied to the 8B PaLM model, with an improvement of +13 points in $F_{0.5}$. This further highlights the ability of \ours{} to compensate for low-quality candidates by merging and correcting them in order to achieve competitive performance. Furthermore, we observe that using a single candidate for \ours{} leads to inferior performance in all cases, except for the 62B PaLM model. This finding suggests that the existence of diverse candidates prevents the model from overfitting to the outputs of a specific LLM, thereby enhancing its generalization capabilities.

\begin{table}[t]
\centering
\resizebox{0.8\columnwidth}{!}{%
\begin{tabular}{lc|c|c}
\hline
T5-base & \multicolumn{3}{c}{59.38}\\
\hline
\multirow{ 2}{*}{PaLM (ICL)}  & \textit{8B} & \textit{62B} & \textit{540B}\\ \cline{2-4}
& 48.62 & 59.92 & \textbf{65.37}  \\ \hline
+ \ours{} (single)  & 61.40 & \textbf{62.48}  & 63.55  \\
+ \ours{} (mult.)  & \textbf{61.89} & \textbf{62.47} & 65.16  \\ \hline
\end{tabular}
}
\caption{Results ($F_{0.5}$) of applying the corrector to LMs of different scale during inference for GEC.}
\label{tab:different_scales}
\end{table}

\begin{table}[t]
\centering
\resizebox{0.67\columnwidth}{!}{%
\begin{tabular}{lcc}
\hline
\textbf{Model} & \textbf{R-2} & \textbf{R-L} \\ \hline
GPT3-Codex (ICL)* & 34.2 & 44.4  \\ 
+ MBRD-BLEURT* & 36.4  & 46.5  \\ \hline
+ \ours{} (mult.) & \textbf{44.8} & \textbf{53.0} \\ \hline
\end{tabular}
}
\caption{Applying the corrector to different family of LMs during inference for E2E NLG. \textit{*}: the results as reported in the original paper.}
\label{tab:different_LMs}
\end{table}

As a next step, we explore the application of \ours{} to an LLM from a distinct family of models, specifically Codex\footnote{We use the outputs of \texttt{code-davinci-002} provided by \citet{mbrd}.} which is a GPT3-like model trained on code~\cite{Chen2021EvaluatingLL}. To compare the effectiveness of \ours{} with MRBD reranking we utilize BLEURT~\cite{sellam-etal-2020-bleurt} as the alignment function, as it has been reported to achieve the highest scores for the E2E NLG task~\cite{suzgun-etal-2022-prompt}. It is important to note that while the reranking approach samples 16 outputs from the LLM, we only use 5 for the corrector. The results in Table~\ref{tab:different_LMs} demonstrate the superior performance of \ours{} over reranking. In particular we observe a performance boost of 10 points in ROUGE-2 when compared to in-context learning, whereas MBRD achieves a mere 2-point improvement.

The previous findings highlight the remarkable out-of-domain robustness of \ours{} and its ability to seamlessly integrate with various LLMs, as a versatile solution for enhancing their performance. This versatility not only holds promise for applying a single corrector to multiple LLMs but also for training correctors with future, more capable LLMs

\renewcommand*{\arraystretch}{1.1}
\begin{table}[t]
\centering
\resizebox{0.8\columnwidth}{!}{%
\begin{tabular}{lcccc}
\hline
\textbf{Model} & \textbf{R-1} & \textbf{R-2} & \textbf{R-L} & \textbf{BLEU} \\ \hline
Pegasus (FT) & 45.48 & \textbf{23.88} & 38.18 & 16.72 \\ \hline
+ \ours{} & \textbf{45.76} & 23.78 & \textbf{38.28} &\textbf{17.00} \\ \hline
\end{tabular}
}
\caption{Applying the corrector to state-of-the-art summarization model. Results are reported on XSum.}
\label{tab:pegasus_corrector}
\end{table}

\subsection{Task-specific models}\label{sec:pegasus}
To further assess the versatility of \ours{} we extend our investigation to specialized models. Specifically, we train a corrector using candidates produced by Pegasus~\cite{pegasus}, a state-of-the-art summarization model, via beam search. The results of Table~\ref{tab:pegasus_corrector} reveal that \ours{} provides gains even for models that have undergone pre-training and fine-tuning tailored to the task. Although the gains are relatively modest compared to PaLM, this discrepancy can be attributed to the already high performance of Pegasus and the lack of diversity of the beam-generated candidates, which is essential for the corrector. The increased performance of the corrector when applied to Pegasus compared to PaLM supports our intuition regarding the noise introduced by low-quality candidates.

\section{Analysis}

\subsection{Importance of the source}
The input of the corrector consists of the source sentence and a number of candidates generated by the LLM (Equation~\ref{eq:corr_input}). In  previous sections we demonstrated that the use of multiple candidates improves in-domain performance and out-of-domain robustness. In this experiment, we focus on the importance of the source sentence to \ours{}. To examine this, we train a corrector that receives only the candidates as input, without access to the source. The results for E2E NLG, presented in Table~\ref{tab:source_corrector} reveal a noticeable decline in performance when the source sentence is removed. This decrease can be attributed to the inability of the corrector to produce outputs that are faithful to the input in the absence of the source sentence. 

\subsection{Scaling the Corrector} \label{sec:model_scaling}
We showed that a corrector with 250M parameters, can effectively refine the predictions of LLMs for specific tasks. This raises the question: \textit{is training a corrector still valuable if we have the computational resources to train a very large model?} To investigate this we train the largest version of T5, T5-xxl with 11B parameters, both through standard fine-tuning and as a corrector for GEC. We note that, in this scenario, the sizes of the fine-tuned model and the LLM are comparable (11B vs. 62B).

\begin{table}[t]
\centering
\resizebox{0.56\columnwidth}{!}{%
\begin{tabular}{lcc}
\hline
\textbf{Model} & \textbf{R-2} & \textbf{R-L} \\ \hline
PaLM-62B (ICL) & 35.1 & 45.6 \\ \hline
+ \ours{} & \textbf{45.6} & \textbf{53.4} \\
- source sentence & 44.5 & 53.1 \\ \hline
\end{tabular}
}
\caption{The importance of the source sentence for the corrector. Results are reported on E2E NLG (clean).}
\label{tab:source_corrector}
\end{table}

\begin{figure}[t]
    \centering
    \includegraphics[width=0.8\columnwidth]{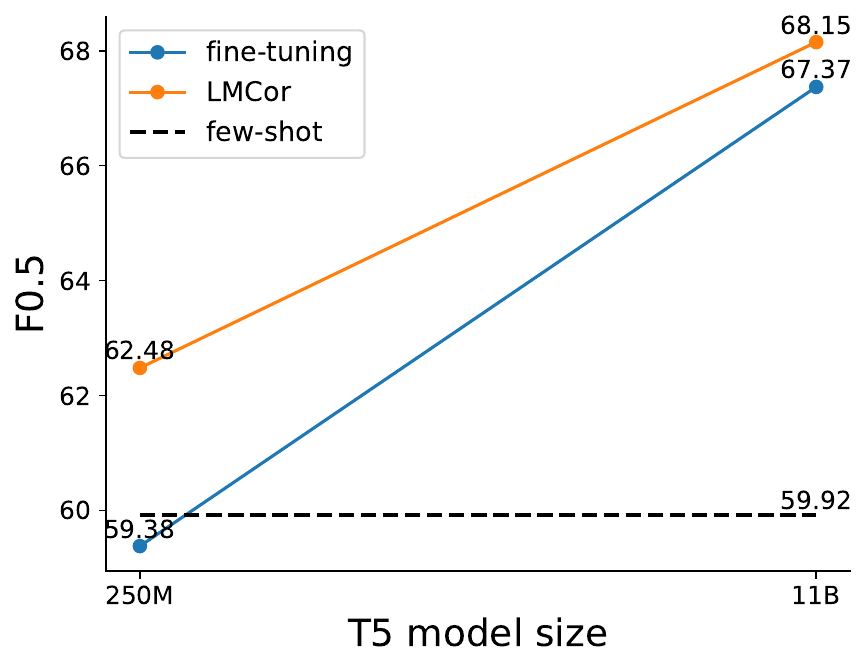}
    \caption{The effect of scaling for \ours{} and fine-tuning. Results are reported on GEC.}
    \label{fig:scaling}
\end{figure}

As shown in Figure~\ref{fig:scaling} both \ours{} and the fine-tuned T5 benefit from scaling, exhibiting higher $F0.5$ scores as their parameter count increases from 250 million to 11 billion. At the 11 billion scale, both models significantly outperfrom in-context learning with the 62B PaLM model. The corrector continues to outperform the baseline as model scale increases, although the gap in performance narrows from 3 to 0.5 points in $F0.5$. We attribute this reduction to the enhanced competence of the larger T5 model. At the scale of 11 billion parameters, the scores obtained by the T5 model alone surpass those obtained by the LLM, indicating that the LLM-generated outputs are of lower quality compared to the ones that the model would produce independently. Therefore, while the additional input to the model remains beneficial, its impact diminishes to some extent, considering the performance disparity between T5 and PaLM.

\section{Related Work}

Since the introduction of in-context learning, prior research has primarily focused on improving the few-shot performance of LLMs. One approach suggests prompting the model to generate rationales or chain-of-thoughts~\cite{scratchpad,wei2022chain, kojima2022large} or to decompose the problem into simpler ones~\cite{press2022measuring, zhou2023leasttomost, pilault2023interactive}, simulating a reasoning process prior to generating the answer. 
These prompting techniques are complementary to our approach and can provide improved candidates for the corrector.

Another strategy to improve the performance of LLMs without training is reranking, i.e. selecting the most promising from a pool of candidates obtained by sampling from the model. Reranking approaches include training different models as the ranker~\cite{gsm8k}, using task-specific ranking functions~\cite{suzgun-etal-2022-prompt, fernandes-etal-2022-quality}, majority voting~\cite{wang2023selfconsistency} or minimum Bayes risk decoding (MBRD)~\cite{mbrd, freitag-etal-2022-high}. Although these approaches can improve the few-shot performance of an LLM, they are upper-bounded by the quality of the generated candidates.

While fine-tuning large LLMs can enhance performance, the substantial computational requirements have prompted the development of parameter-efficient fine-tuning methods (PEFT) \cite{he2022towards}. These approaches introduce a small number of additional parameters (relative to the full model) to be trained while the rest of the model is frozen. The newly-added parameters can come in the form of embeddings that are appended to the encoded sequence~\cite{li-liang-2021-prefix, lester-etal-2021-power}, MLPs that are added in-between layers, namely adapters~\cite{pmlr-v97-houlsby19a, karimi2021compacter} or rank-decomposition matrices that are added in parallel to the existing layers~\cite{hu2022lora, Zhang2023LLaMAAdapterEF}. 
Although these works decrease the computational load of fine-tuning they still require loading and backpropagating through the model, which can be prohibitive for LLMs. Our work shares the same motivation with PEFT methods, with the introduced parameters being essentially another, smaller model. However, our method does not have the memory requirements of PEFT since the corrector operates directly on the model's outputs and does not require access to the model's weights.


An alternative line of work proposes providing feedback to the LLM in order to revise and enhance its predictions. The feedback can be obtained from external models 
such as Google Search, document retrievers, compilers~\citep{gao-etal-2021-making, yao2023react, peng2023check, gou2023critic}, or from a separate model trained to provide feedback on LLM outputs with additional supervision~\citep{paul2023refiner, peng2023check, rl4rf}. While leveraging the LLM itself to generate feedback has been explored~\citep{madaan2023selfrefine, shinn2023reflexion}, it tends to yield lower-quality feedback~\citep{rl4rf, gou2023critic, huang2023large} and involves multiple passes and extensive prompt engineering for each LLM operation. In contrast, our approach is task-agnostic and requires a single pass from the LLM, with little to no prompt engineering, offering an efficient solution for enhancing LLM outputs.


Recently, studies have highlighted the potential of smaller, task-specific models to complement the predictions of an LLM. \citet{xu2023small} explore a framework where candidates produced by task-specific models are fed to an LLM, primarily targeting classification task while \ours{} is better-suited for open-ended generation tasks. \citet{welleck2023generating} train a smaller model to iteratively improve sequences generated by LLMs. In contrast to our method, they rely on unlabeled data and sample extensively from the LLM to obtain a large pool of candidates. They assume the availability of a value function that assigns scores to each candidate and create input-output pairs by sorting candidates based on their scores. 
Unlike their approach, we demonstrate that a compact corrector can perform effectively across various tasks. Additionally, our approach is more efficient during inference, since the ability of \ours{} to process multiple candidates simultaneously eliminates the need for multiple passes. 

Concurrently, researchers have begun to leverage the complementary nature of LLM-generated outputs during inference. \citet{farinhas-etal-2023-empirical} use an LLM to combine its generated outputs for machine translation, although they find that reranking methods incorporating external modules, such as quality estimation metrics~\cite{zerva-etal-2022-findings}, prove to be more efficient. Meanwhile, \citet{vernikos2024dont} propose an approach that uses a quality estimation metric to combine the outputs of LLMs or MT models. Similar to our method, they exploit the diversity of LLM outputs by identifying divergent spans among candidates and merging them based on the metric.

Most relevant to our approach is the work by \citet{jiang-etal-2023-llm} where they propose a method to ensemble LLMs. Their pipeline consists of i) sampling a large pool of candidates, ii) selecting top candidates via multiple pairwise comparisons through a trained reranker and iii) fusing them using a similar technique as \ours{}. While our approach could be extended to multiple LLMs we demonstrate improvements with a single LLM, leveraging the complimentarity of the generations. In addition, our approach is more efficient since we use a much smaller model as the corrector (3B vs 250M) and do not introduce additional training and inference steps for ranking the outputs. 


\section{Conclusion}
In this work, we introduce \ours{}, a novel approach that leverages a small corrector module to enhance the performance of LLMs in the presence of training data. \ours{} leverages the diversity of the LLM generations to rank, edit and combine the candidates. Unlike parameter-efficient fine-tuning methods, our approach does not require access to the model or substantial computational resources. Our experiments demonstrate that even a relatively small corrector (250M) can improve the performance of a much larger LM (62B), while exhibiting robustness against different prompts. Furthermore, we showcase that the corrector can be successfully applied to models of different scale or architecture without any retraining. These findings offer a promising solution for improving LLM performance in a practical and resource-efficient manner and open up new possibilities for the utilization and deployment of LLMs in real-world applications alongside smaller task-specific models.

\section*{Acknowledgments}

We are grateful for their support to the Swiss National Science Foundation (DOMAT grant n.\ 175693, On-demand Knowledge for Document-level Machine Translation), and to the Institute for ICT at HEIG-VD. We thank Andrei Popescu-Belis and Katerina Margatina for their valuable comments and fruitful discussions.

\section*{Limitations}
\paragraph{Additional latency} While our approach enhances the performance of LLMs on the considered tasks, it also introduces additional latency. Instead of a single pass from the LLM our pipeline involves sampling multiple candidates in parallel and performing an additional inference step using a much smaller model. Although the additional latency is small, it could be critical for low-latency applications that require real-time responses.

\paragraph{Strong fine-tuned models} While our approach demonstrates gains over in-context learning and reranking it may not always achieve the same level of performance as fine-tuning approaches. Our results in XSum and Figure~\ref{fig:scaling_data} indicate that fine-tuning remains a powerful method for smaller models when ample data is available. Additionally, scaling fine-tuned models instead of using an off-the-shelf LLM might be a better alternative in certain cases, as discussed in Section~\ref{sec:model_scaling}.

\paragraph{Additional tasks and models} Due to time and budget limit our experiments cover 4 natural language generation tasks and they could be extended to other kinds of tasks such as reasoning. Additionally while \ours{} shows promising results when combined with different LLMs, even during inference, it would be interesting to apply our approach to a  broader selection of LLMs that are currently available.

\paragraph{Human evaluation} We agree that automatic metrics have limitations. While the selection of metrics aligns with prior work, human evaluation could provide us with more reliable and comprehensive evaluation results. However, due to the number of models and the amount of generation candidates, we could not afford large-scale human evaluation.

\section*{Broader Impact}
Given that the proposed approach combines LLMs and small corrector models for improved performance, it is important to acknowledge that it shares the potential social biases associated with LLMs. While our work focuses on improving the predictions of LLMs on specific datasets rather than open-ended generation, it is improbable that our approach amplifies these biases to a greater extent than other methods. Nonetheless, it is important to investigate whether \ours{} has any impact on either amplifying or mitigating these biases.

\bibliography{anthology,custom}

\begin{thebibliography}{64}
\expandafter\ifx\csname natexlab\endcsname\relax\def\natexlab#1{#1}\fi

\bibitem[{Akhbardeh et~al.(2021)Akhbardeh, Arkhangorodsky, Biesialska, Bojar, Chatterjee, Chaudhary, Costa-jussa, Espa{\~n}a-Bonet, Fan, Federmann, Freitag, Graham, Grundkiewicz, Haddow, Harter, Heafield, Homan, Huck, Amponsah-Kaakyire, Kasai, Khashabi, Knight, Kocmi, Koehn, Lourie, Monz, Morishita, Nagata, Nagesh, Nakazawa, Negri, Pal, Tapo, Turchi, Vydrin, and Zampieri}]{akhbardeh-etal-2021-findings}
Farhad Akhbardeh, Arkady Arkhangorodsky, Magdalena Biesialska, Ond{\v{r}}ej Bojar, Rajen Chatterjee, Vishrav Chaudhary, Marta~R. Costa-jussa, Cristina Espa{\~n}a-Bonet, Angela Fan, Christian Federmann, Markus Freitag, Yvette Graham, Roman Grundkiewicz, Barry Haddow, Leonie Harter, Kenneth Heafield, Christopher Homan, Matthias Huck, Kwabena Amponsah-Kaakyire, Jungo Kasai, Daniel Khashabi, Kevin Knight, Tom Kocmi, Philipp Koehn, Nicholas Lourie, Christof Monz, Makoto Morishita, Masaaki Nagata, Ajay Nagesh, Toshiaki Nakazawa, Matteo Negri, Santanu Pal, Allahsera~Auguste Tapo, Marco Turchi, Valentin Vydrin, and Marcos Zampieri. 2021.
\newblock \href {https://aclanthology.org/2021.wmt-1.1} {Findings of the 2021 conference on machine translation ({WMT}21)}.
\newblock In \emph{Proceedings of the Sixth Conference on Machine Translation}, pages 1--88.

\bibitem[{Akyürek et~al.(2023)Akyürek, Akyürek, Madaan, Kalyan, Clark, Wijaya, and Tandon}]{rl4rf}
Afra~Feyza Akyürek, Ekin Akyürek, Aman Madaan, Ashwin Kalyan, Peter Clark, Derry Wijaya, and Niket Tandon. 2023.
\newblock \href {http://arxiv.org/abs/2305.08844} {Rl4f: Generating natural language feedback with reinforcement learning for repairing model outputs}.

\bibitem[{Brown et~al.(2020{\natexlab{a}})Brown, Mann, Ryder, Subbiah, Kaplan, Dhariwal, Neelakantan, Shyam, Sastry, Askell, Agarwal, Herbert-Voss, Krueger, Henighan, Child, Ramesh, Ziegler, Wu, Winter, Hesse, Chen, Sigler, Litwin, Gray, Chess, Clark, Berner, McCandlish, Radford, Sutskever, and Amodei}]{GPT3}
Tom Brown, Benjamin Mann, Nick Ryder, Melanie Subbiah, Jared~D Kaplan, Prafulla Dhariwal, Arvind Neelakantan, Pranav Shyam, Girish Sastry, Amanda Askell, Sandhini Agarwal, Ariel Herbert-Voss, Gretchen Krueger, Tom Henighan, Rewon Child, Aditya Ramesh, Daniel Ziegler, Jeffrey Wu, Clemens Winter, Chris Hesse, Mark Chen, Eric Sigler, Mateusz Litwin, Scott Gray, Benjamin Chess, Jack Clark, Christopher Berner, Sam McCandlish, Alec Radford, Ilya Sutskever, and Dario Amodei. 2020{\natexlab{a}}.
\newblock \href {https://proceedings.neurips.cc/paper/2020/file/1457c0d6bfcb4967418bfb8ac142f64a-Paper.pdf} {Language models are few-shot learners}.
\newblock In \emph{Advances in Neural Information Processing Systems}, volume~33, pages 1877--1901.

\bibitem[{Brown et~al.(2020{\natexlab{b}})Brown, Mann, Ryder, Subbiah, Kaplan, Dhariwal, Neelakantan, Shyam, Sastry, Askell et~al.}]{brown2020language}
Tom Brown, Benjamin Mann, Nick Ryder, Melanie Subbiah, Jared~D Kaplan, Prafulla Dhariwal, Arvind Neelakantan, Pranav Shyam, Girish Sastry, Amanda Askell, et~al. 2020{\natexlab{b}}.
\newblock Language models are few-shot learners.
\newblock \emph{Advances in neural information processing systems}, 33:1877--1901.

\bibitem[{Bryant et~al.(2019)Bryant, Felice, Andersen, and Briscoe}]{bryant-etal-2019-bea}
Christopher Bryant, Mariano Felice, {\O}istein~E. Andersen, and Ted Briscoe. 2019.
\newblock \href {https://doi.org/10.18653/v1/W19-4406} {The {BEA}-2019 shared task on grammatical error correction}.
\newblock In \emph{Proceedings of the Fourteenth Workshop on Innovative Use of NLP for Building Educational Applications}, pages 52--75.

\bibitem[{Chen et~al.(2021)Chen, Tworek, Jun, Yuan, Ponde, Kaplan, Edwards, Burda, Joseph, Brockman, Ray, Puri, Krueger, Petrov, Khlaaf, Sastry, Mishkin, Chan, Gray, Ryder, Pavlov, Power, Kaiser, Bavarian, Winter, Tillet, Such, Cummings, Plappert, Chantzis, Barnes, Herbert-Voss, Guss, Nichol, Babuschkin, Balaji, Jain, Carr, Leike, Achiam, Misra, Morikawa, Radford, Knight, Brundage, Murati, Mayer, Welinder, McGrew, Amodei, McCandlish, Sutskever, and Zaremba}]{Chen2021EvaluatingLL}
Mark Chen, Jerry Tworek, Heewoo Jun, Qiming Yuan, Henrique Ponde, Jared Kaplan, Harrison Edwards, Yura Burda, Nicholas Joseph, Greg Brockman, Alex Ray, Raul Puri, Gretchen Krueger, Michael Petrov, Heidy Khlaaf, Girish Sastry, Pamela Mishkin, Brooke Chan, Scott Gray, Nick Ryder, Mikhail Pavlov, Alethea Power, Lukasz Kaiser, Mohammad Bavarian, Clemens Winter, Philippe Tillet, Felipe~Petroski Such, David~W. Cummings, Matthias Plappert, Fotios Chantzis, Elizabeth Barnes, Ariel Herbert-Voss, William~H. Guss, Alex Nichol, Igor Babuschkin, S.~Arun Balaji, Shantanu Jain, Andrew Carr, Jan Leike, Joshua Achiam, Vedant Misra, Evan Morikawa, Alec Radford, Matthew~M. Knight, Miles Brundage, Mira Murati, Katie Mayer, Peter Welinder, Bob McGrew, Dario Amodei, Sam McCandlish, Ilya Sutskever, and Wojciech Zaremba. 2021.
\newblock Evaluating large language models trained on code.
\newblock \emph{ArXiv}, abs/2107.03374.

\bibitem[{Chowdhery et~al.(2022)Chowdhery, Narang, Devlin, Bosma, Mishra, Roberts, Barham, Chung, Sutton, Gehrmann, Schuh, Shi, Tsvyashchenko, Maynez, Rao, Barnes, Tay, Shazeer, Prabhakaran, Reif, Du, Hutchinson, Pope, Bradbury, Austin, Isard, Gur-Ari, Yin, Duke, Levskaya, Ghemawat, Dev, Michalewski, Garcia, Misra, Robinson, Fedus, Zhou, Ippolito, Luan, Lim, Zoph, Spiridonov, Sepassi, Dohan, Agrawal, Omernick, Dai, Pillai, Pellat, Lewkowycz, Moreira, Child, Polozov, Lee, Zhou, Wang, Saeta, Diaz, Firat, Catasta, Wei, Meier-Hellstern, Eck, Dean, Petrov, and Fiedel}]{palm}
Aakanksha Chowdhery, Sharan Narang, Jacob Devlin, Maarten Bosma, Gaurav Mishra, Adam Roberts, Paul Barham, Hyung~Won Chung, Charles Sutton, Sebastian Gehrmann, Parker Schuh, Kensen Shi, Sasha Tsvyashchenko, Joshua Maynez, Abhishek Rao, Parker Barnes, Yi~Tay, Noam Shazeer, Vinodkumar Prabhakaran, Emily Reif, Nan Du, Ben Hutchinson, Reiner Pope, James Bradbury, Jacob Austin, Michael Isard, Guy Gur-Ari, Pengcheng Yin, Toju Duke, Anselm Levskaya, Sanjay Ghemawat, Sunipa Dev, Henryk Michalewski, Xavier Garcia, Vedant Misra, Kevin Robinson, Liam Fedus, Denny Zhou, Daphne Ippolito, David Luan, Hyeontaek Lim, Barret Zoph, Alexander Spiridonov, Ryan Sepassi, David Dohan, Shivani Agrawal, Mark Omernick, Andrew~M. Dai, Thanumalayan~Sankaranarayana Pillai, Marie Pellat, Aitor Lewkowycz, Erica Moreira, Rewon Child, Oleksandr Polozov, Katherine Lee, Zongwei Zhou, Xuezhi Wang, Brennan Saeta, Mark Diaz, Orhan Firat, Michele Catasta, Jason Wei, Kathy Meier-Hellstern, Douglas Eck, Jeff Dean, Slav Petrov, and Noah Fiedel. 2022.
\newblock \href {https://doi.org/10.48550/ARXIV.2204.02311} {Palm: Scaling language modeling with pathways}.

\bibitem[{Cobbe et~al.(2021)Cobbe, Kosaraju, Bavarian, Hilton, Nakano, Hesse, and Schulman}]{gsm8k}
Karl Cobbe, Vineet Kosaraju, Mohammad Bavarian, Jacob Hilton, Reiichiro Nakano, Christopher Hesse, and John Schulman. 2021.
\newblock \href {http://arxiv.org/abs/2110.14168} {Training verifiers to solve math word problems}.
\newblock \emph{CoRR}, abs/2110.14168.

\bibitem[{Dahlmeier and Ng(2012)}]{dahlmeier-ng-2012-better}
Daniel Dahlmeier and Hwee~Tou Ng. 2012.
\newblock \href {https://aclanthology.org/N12-1067} {Better evaluation for grammatical error correction}.
\newblock In \emph{Proceedings of the 2012 Conference of the North {A}merican Chapter of the Association for Computational Linguistics: Human Language Technologies}, pages 568--572.

\bibitem[{Du{\v{s}}ek et~al.(2019)Du{\v{s}}ek, Howcroft, and Rieser}]{dusek-etal-2019-semantic}
Ond{\v{r}}ej Du{\v{s}}ek, David~M. Howcroft, and Verena Rieser. 2019.
\newblock \href {https://doi.org/10.18653/v1/W19-8652} {Semantic noise matters for neural natural language generation}.
\newblock In \emph{Proceedings of the 12th International Conference on Natural Language Generation}, pages 421--426.

\bibitem[{Farinhas et~al.(2023)Farinhas, de~Souza, and Martins}]{farinhas-etal-2023-empirical}
Ant{\'o}nio Farinhas, Jos{\'e} de~Souza, and Andre Martins. 2023.
\newblock \href {https://doi.org/10.18653/v1/2023.emnlp-main.733} {An empirical study of translation hypothesis ensembling with large language models}.
\newblock In \emph{Proceedings of the 2023 Conference on Empirical Methods in Natural Language Processing}, pages 11956--11970.

\bibitem[{Fernandes et~al.(2022)Fernandes, Farinhas, Rei, C.~de Souza, Ogayo, Neubig, and Martins}]{fernandes-etal-2022-quality}
Patrick Fernandes, Ant{\'o}nio Farinhas, Ricardo Rei, Jos{\'e}~G. C.~de Souza, Perez Ogayo, Graham Neubig, and Andre Martins. 2022.
\newblock \href {https://doi.org/10.18653/v1/2022.naacl-main.100} {Quality-aware decoding for neural machine translation}.
\newblock In \emph{Proceedings of the 2022 Conference of the North American Chapter of the Association for Computational Linguistics: Human Language Technologies}, pages 1396--1412.

\bibitem[{Freitag et~al.(2022)Freitag, Grangier, Tan, and Liang}]{freitag-etal-2022-high}
Markus Freitag, David Grangier, Qijun Tan, and Bowen Liang. 2022.
\newblock \href {https://doi.org/10.1162/tacl_a_00491} {High quality rather than high model probability: Minimum {B}ayes risk decoding with neural metrics}.
\newblock \emph{Transactions of the Association for Computational Linguistics}, 10:811--825.

\bibitem[{Fu et~al.(2023)Fu, Peng, Ou, Sabharwal, and Khot}]{fu2023specializing}
Yao Fu, Hao Peng, Litu Ou, Ashish Sabharwal, and Tushar Khot. 2023.
\newblock \href {https://proceedings.mlr.press/v202/fu23d.html} {Specializing smaller language models towards multi-step reasoning}.
\newblock In \emph{Proceedings of the 40th International Conference on Machine Learning}, volume 202 of \emph{Proceedings of Machine Learning Research}, pages 10421--10430.

\bibitem[{Gao et~al.(2021)Gao, Fisch, and Chen}]{gao-etal-2021-making}
Tianyu Gao, Adam Fisch, and Danqi Chen. 2021.
\newblock \href {https://doi.org/10.18653/v1/2021.acl-long.295} {Making pre-trained language models better few-shot learners}.
\newblock In \emph{Proceedings of the 59th Annual Meeting of the Association for Computational Linguistics and the 11th International Joint Conference on Natural Language Processing (Volume 1: Long Papers)}, pages 3816--3830.

\bibitem[{Gou et~al.(2023)Gou, Shao, Gong, Shen, Yang, Duan, and Chen}]{gou2023critic}
Zhibin Gou, Zhihong Shao, Yeyun Gong, Yelong Shen, Yujiu Yang, Nan Duan, and Weizhu Chen. 2023.
\newblock \href {http://arxiv.org/abs/2305.11738} {Critic: Large language models can self-correct with tool-interactive critiquing}.

\bibitem[{He et~al.(2022)He, Zhou, Ma, Berg-Kirkpatrick, and Neubig}]{he2022towards}
Junxian He, Chunting Zhou, Xuezhe Ma, Taylor Berg-Kirkpatrick, and Graham Neubig. 2022.
\newblock \href {https://openreview.net/forum?id=0RDcd5Axok} {Towards a unified view of parameter-efficient transfer learning}.
\newblock In \emph{International Conference on Learning Representations}.

\bibitem[{Holtzman et~al.(2020)Holtzman, Buys, Du, Forbes, and Choi}]{Holtzman2020The}
Ari Holtzman, Jan Buys, Li~Du, Maxwell Forbes, and Yejin Choi. 2020.
\newblock \href {https://openreview.net/forum?id=rygGQyrFvH} {The curious case of neural text degeneration}.
\newblock In \emph{International Conference on Learning Representations}.

\bibitem[{Houlsby et~al.(2019)Houlsby, Giurgiu, Jastrzebski, Morrone, De~Laroussilhe, Gesmundo, Attariyan, and Gelly}]{pmlr-v97-houlsby19a}
Neil Houlsby, Andrei Giurgiu, Stanislaw Jastrzebski, Bruna Morrone, Quentin De~Laroussilhe, Andrea Gesmundo, Mona Attariyan, and Sylvain Gelly. 2019.
\newblock \href {https://proceedings.mlr.press/v97/houlsby19a.html} {Parameter-efficient transfer learning for {NLP}}.
\newblock In \emph{Proceedings of the 36th International Conference on Machine Learning}, volume~97 of \emph{Proceedings of Machine Learning Research}, pages 2790--2799.

\bibitem[{Hu et~al.(2022)Hu, yelong shen, Wallis, Allen-Zhu, Li, Wang, Wang, and Chen}]{hu2022lora}
Edward~J Hu, yelong shen, Phillip Wallis, Zeyuan Allen-Zhu, Yuanzhi Li, Shean Wang, Lu~Wang, and Weizhu Chen. 2022.
\newblock \href {https://openreview.net/forum?id=nZeVKeeFYf9} {Lo{RA}: Low-rank adaptation of large language models}.
\newblock In \emph{International Conference on Learning Representations}.

\bibitem[{Huang et~al.(2023)Huang, Chen, Mishra, Zheng, Yu, Song, and Zhou}]{huang2023large}
Jie Huang, Xinyun Chen, Swaroop Mishra, Huaixiu~Steven Zheng, Adams~Wei Yu, Xinying Song, and Denny Zhou. 2023.
\newblock \href {http://arxiv.org/abs/2310.01798} {Large language models cannot self-correct reasoning yet}.

\bibitem[{Jiang et~al.(2023)Jiang, Ren, and Lin}]{jiang-etal-2023-llm}
Dongfu Jiang, Xiang Ren, and Bill~Yuchen Lin. 2023.
\newblock \href {https://doi.org/10.18653/v1/2023.acl-long.792} {{LLM}-blender: Ensembling large language models with pairwise ranking and generative fusion}.
\newblock In \emph{Proceedings of the 61st Annual Meeting of the Association for Computational Linguistics (Volume 1: Long Papers)}, pages 14165--14178.

\bibitem[{Jiang et~al.(2021)Jiang, Araki, Ding, and Neubig}]{jiang-etal-2021-know}
Zhengbao Jiang, Jun Araki, Haibo Ding, and Graham Neubig. 2021.
\newblock \href {https://doi.org/10.1162/tacl_a_00407} {How can we know when language models know? on the calibration of language models for question answering}.
\newblock \emph{Transactions of the Association for Computational Linguistics}, 9:962--977.

\bibitem[{Karimi~Mahabadi et~al.(2021)Karimi~Mahabadi, Henderson, and Ruder}]{karimi2021compacter}
Rabeeh Karimi~Mahabadi, James Henderson, and Sebastian Ruder. 2021.
\newblock Compacter: Efficient low-rank hypercomplex adapter layers.
\newblock \emph{Advances in Neural Information Processing Systems}, 34:1022--1035.

\bibitem[{Kocmi et~al.(2022)Kocmi, Bawden, Bojar, Dvorkovich, Federmann, Fishel, Gowda, Graham, Grundkiewicz, Haddow, Knowles, Koehn, Monz, Morishita, Nagata, Nakazawa, Nov{\'a}k, Popel, and Popovi{\'c}}]{kocmi-etal-2022-findings}
Tom Kocmi, Rachel Bawden, Ond{\v{r}}ej Bojar, Anton Dvorkovich, Christian Federmann, Mark Fishel, Thamme Gowda, Yvette Graham, Roman Grundkiewicz, Barry Haddow, Rebecca Knowles, Philipp Koehn, Christof Monz, Makoto Morishita, Masaaki Nagata, Toshiaki Nakazawa, Michal Nov{\'a}k, Martin Popel, and Maja Popovi{\'c}. 2022.
\newblock \href {https://aclanthology.org/2022.wmt-1.1} {Findings of the 2022 conference on machine translation ({WMT}22)}.
\newblock In \emph{Proceedings of the Seventh Conference on Machine Translation (WMT)}, pages 1--45.

\bibitem[{Kojima et~al.(2022)Kojima, Gu, Reid, Matsuo, and Iwasawa}]{kojima2022large}
Takeshi Kojima, Shixiang~Shane Gu, Machel Reid, Yutaka Matsuo, and Yusuke Iwasawa. 2022.
\newblock \href {https://openreview.net/forum?id=6p3AuaHAFiN} {Large language models are zero-shot reasoners}.
\newblock In \emph{ICML 2022 Workshop on Knowledge Retrieval and Language Models}.

\bibitem[{Lester et~al.(2021)Lester, Al-Rfou, and Constant}]{lester-etal-2021-power}
Brian Lester, Rami Al-Rfou, and Noah Constant. 2021.
\newblock \href {https://doi.org/10.18653/v1/2021.emnlp-main.243} {The power of scale for parameter-efficient prompt tuning}.
\newblock In \emph{Proceedings of the 2021 Conference on Empirical Methods in Natural Language Processing}, pages 3045--3059.

\bibitem[{Li and Liang(2021)}]{li-liang-2021-prefix}
Xiang~Lisa Li and Percy Liang. 2021.
\newblock \href {https://doi.org/10.18653/v1/2021.acl-long.353} {Prefix-tuning: Optimizing continuous prompts for generation}.
\newblock In \emph{Proceedings of the 59th Annual Meeting of the Association for Computational Linguistics and the 11th International Joint Conference on Natural Language Processing (Volume 1: Long Papers)}, pages 4582--4597.

\bibitem[{Lin(2004)}]{lin-2004-rouge}
Chin-Yew Lin. 2004.
\newblock \href {https://aclanthology.org/W04-1013} {{ROUGE}: A package for automatic evaluation of summaries}.
\newblock In \emph{Text Summarization Branches Out}, pages 74--81.

\bibitem[{Lin et~al.(2022)Lin, Mihaylov, Artetxe, Wang, Chen, Simig, Ott, Goyal, Bhosale, Du, Pasunuru, Shleifer, Koura, Chaudhary, O{'}Horo, Wang, Zettlemoyer, Kozareva, Diab, Stoyanov, and Li}]{lin-etal-2022-shot}
Xi~Victoria Lin, Todor Mihaylov, Mikel Artetxe, Tianlu Wang, Shuohui Chen, Daniel Simig, Myle Ott, Naman Goyal, Shruti Bhosale, Jingfei Du, Ramakanth Pasunuru, Sam Shleifer, Punit~Singh Koura, Vishrav Chaudhary, Brian O{'}Horo, Jeff Wang, Luke Zettlemoyer, Zornitsa Kozareva, Mona Diab, Veselin Stoyanov, and Xian Li. 2022.
\newblock \href {https://doi.org/10.18653/v1/2022.emnlp-main.616} {Few-shot learning with multilingual generative language models}.
\newblock In \emph{Proceedings of the 2022 Conference on Empirical Methods in Natural Language Processing}, pages 9019--9052.

\bibitem[{Liu et~al.(2022)Liu, Shen, Zhang, Dolan, Carin, and Chen}]{liu-etal-2022-makes}
Jiachang Liu, Dinghan Shen, Yizhe Zhang, Bill Dolan, Lawrence Carin, and Weizhu Chen. 2022.
\newblock \href {https://doi.org/10.18653/v1/2022.deelio-1.10} {What makes good in-context examples for {GPT}-3?}
\newblock In \emph{Proceedings of Deep Learning Inside Out (DeeLIO 2022): The 3rd Workshop on Knowledge Extraction and Integration for Deep Learning Architectures}, pages 100--114.

\bibitem[{Lu et~al.(2022)Lu, Bartolo, Moore, Riedel, and Stenetorp}]{lu-etal-2022-fantastically}
Yao Lu, Max Bartolo, Alastair Moore, Sebastian Riedel, and Pontus Stenetorp. 2022.
\newblock \href {https://doi.org/10.18653/v1/2022.acl-long.556} {Fantastically ordered prompts and where to find them: Overcoming few-shot prompt order sensitivity}.
\newblock In \emph{Proceedings of the 60th Annual Meeting of the Association for Computational Linguistics (Volume 1: Long Papers)}, pages 8086--8098.

\bibitem[{Madaan et~al.(2023)Madaan, Tandon, Gupta, Hallinan, Gao, Wiegreffe, Alon, Dziri, Prabhumoye, Yang, Welleck, Majumder, Gupta, Yazdanbakhsh, and Clark}]{madaan2023selfrefine}
Aman Madaan, Niket Tandon, Prakhar Gupta, Skyler Hallinan, Luyu Gao, Sarah Wiegreffe, Uri Alon, Nouha Dziri, Shrimai Prabhumoye, Yiming Yang, Sean Welleck, Bodhisattwa~Prasad Majumder, Shashank Gupta, Amir Yazdanbakhsh, and Peter Clark. 2023.
\newblock \href {http://arxiv.org/abs/2303.17651} {Self-refine: Iterative refinement with self-feedback}.

\bibitem[{Narayan et~al.(2018)Narayan, Cohen, and Lapata}]{narayan2018don}
Shashi Narayan, Shay~B Cohen, and Mirella Lapata. 2018.
\newblock Don't give me the details, just the summary! topic-aware convolutional neural networks for extreme summarization.
\newblock \emph{arXiv preprint arXiv:1808.08745}.

\bibitem[{Ng et~al.(2014)Ng, Wu, Briscoe, Hadiwinoto, Susanto, and Bryant}]{ng-etal-2014-conll}
Hwee~Tou Ng, Siew~Mei Wu, Ted Briscoe, Christian Hadiwinoto, Raymond~Hendy Susanto, and Christopher Bryant. 2014.
\newblock \href {https://doi.org/10.3115/v1/W14-1701} {The {C}o{NLL}-2014 shared task on grammatical error correction}.
\newblock In \emph{Proceedings of the Eighteenth Conference on Computational Natural Language Learning: Shared Task}, pages 1--14.

\bibitem[{Novikova et~al.(2017)Novikova, Du{\v{s}}ek, and Rieser}]{novikova-etal-2017-e2e}
Jekaterina Novikova, Ond{\v{r}}ej Du{\v{s}}ek, and Verena Rieser. 2017.
\newblock \href {https://doi.org/10.18653/v1/W17-5525} {The {E}2{E} dataset: New challenges for end-to-end generation}.
\newblock In \emph{Proceedings of the 18th Annual {SIG}dial Meeting on Discourse and Dialogue}, pages 201--206.

\bibitem[{Nye et~al.(2021)Nye, Andreassen, Gur{-}Ari, Michalewski, Austin, Bieber, Dohan, Lewkowycz, Bosma, Luan, Sutton, and Odena}]{scratchpad}
Maxwell~I. Nye, Anders~Johan Andreassen, Guy Gur{-}Ari, Henryk Michalewski, Jacob Austin, David Bieber, David Dohan, Aitor Lewkowycz, Maarten Bosma, David Luan, Charles Sutton, and Augustus Odena. 2021.
\newblock \href {http://arxiv.org/abs/2112.00114} {Show your work: Scratchpads for intermediate computation with language models}.
\newblock \emph{CoRR}, abs/2112.00114.

\bibitem[{Papineni et~al.(2002)Papineni, Roukos, Ward, and Zhu}]{papineni-etal-2002-bleu}
Kishore Papineni, Salim Roukos, Todd Ward, and Wei-Jing Zhu. 2002.
\newblock \href {https://doi.org/10.3115/1073083.1073135} {{B}leu: a method for automatic evaluation of machine translation}.
\newblock In \emph{Proceedings of the 40th Annual Meeting of the Association for Computational Linguistics}, pages 311--318.

\bibitem[{Paul et~al.(2023)Paul, Ismayilzada, Peyrard, Borges, Bosselut, West, and Faltings}]{paul2023refiner}
Debjit Paul, Mete Ismayilzada, Maxime Peyrard, Beatriz Borges, Antoine Bosselut, Robert West, and Boi Faltings. 2023.
\newblock \href {http://arxiv.org/abs/2304.01904} {Refiner: Reasoning feedback on intermediate representations}.

\bibitem[{Peng et~al.(2023)Peng, Galley, He, Cheng, Xie, Hu, Huang, Liden, Yu, Chen, and Gao}]{peng2023check}
Baolin Peng, Michel Galley, Pengcheng He, Hao Cheng, Yujia Xie, Yu~Hu, Qiuyuan Huang, Lars Liden, Zhou Yu, Weizhu Chen, and Jianfeng Gao. 2023.
\newblock \href {http://arxiv.org/abs/2302.12813} {Check your facts and try again: Improving large language models with external knowledge and automated feedback}.

\bibitem[{Pilault et~al.(2023)Pilault, Garcia, Bra{\v{z}}inskas, and Firat}]{pilault2023interactive}
Jonathan Pilault, Xavier Garcia, Arthur Bra{\v{z}}inskas, and Orhan Firat. 2023.
\newblock Interactive-chain-prompting: Ambiguity resolution for crosslingual conditional generation with interaction.
\newblock \emph{arXiv preprint arXiv:2301.10309}.

\bibitem[{Press et~al.(2022)Press, Zhang, Min, Schmidt, Smith, and Lewis}]{press2022measuring}
Ofir Press, Muru Zhang, Sewon Min, Ludwig Schmidt, Noah~A. Smith, and Mike Lewis. 2022.
\newblock \href {http://arxiv.org/abs/2210.03350} {Measuring and narrowing the compositionality gap in language models}.

\bibitem[{Raffel et~al.(2020)Raffel, Shazeer, Roberts, Lee, Narang, Matena, Zhou, Li, and Liu}]{t5}
Colin Raffel, Noam Shazeer, Adam Roberts, Katherine Lee, Sharan Narang, Michael Matena, Yanqi Zhou, Wei Li, and Peter~J. Liu. 2020.
\newblock Exploring the limits of transfer learning with a unified text-to-text transformer.
\newblock \emph{J. Mach. Learn. Res.}, 21(1).

\bibitem[{Rei et~al.(2022)Rei, C.~de Souza, Alves, Zerva, Farinha, Glushkova, Lavie, Coheur, and Martins}]{rei-etal-2022-comet}
Ricardo Rei, Jos{\'e}~G. C.~de Souza, Duarte Alves, Chrysoula Zerva, Ana~C Farinha, Taisiya Glushkova, Alon Lavie, Luisa Coheur, and Andr{\'e} F.~T. Martins. 2022.
\newblock \href {https://aclanthology.org/2022.wmt-1.52} {{COMET}-22: Unbabel-{IST} 2022 submission for the metrics shared task}.
\newblock In \emph{Proceedings of the Seventh Conference on Machine Translation (WMT)}, pages 578--585.

\bibitem[{Rothe et~al.(2021)Rothe, Mallinson, Malmi, Krause, and Severyn}]{rothe-etal-2021-simple}
Sascha Rothe, Jonathan Mallinson, Eric Malmi, Sebastian Krause, and Aliaksei Severyn. 2021.
\newblock \href {https://doi.org/10.18653/v1/2021.acl-short.89} {A simple recipe for multilingual grammatical error correction}.
\newblock In \emph{Proceedings of the 59th Annual Meeting of the Association for Computational Linguistics and the 11th International Joint Conference on Natural Language Processing (Volume 2: Short Papers)}, pages 702--707.

\bibitem[{Schick and Sch{\"u}tze(2021)}]{schick-schutze-2021-just}
Timo Schick and Hinrich Sch{\"u}tze. 2021.
\newblock \href {https://doi.org/10.18653/v1/2021.naacl-main.185} {It{'}s not just size that matters: Small language models are also few-shot learners}.
\newblock In \emph{Proceedings of the 2021 Conference of the North American Chapter of the Association for Computational Linguistics: Human Language Technologies}, pages 2339--2352.

\bibitem[{Sellam et~al.(2020)Sellam, Das, and Parikh}]{sellam-etal-2020-bleurt}
Thibault Sellam, Dipanjan Das, and Ankur Parikh. 2020.
\newblock \href {https://doi.org/10.18653/v1/2020.acl-main.704} {{BLEURT}: Learning robust metrics for text generation}.
\newblock In \emph{Proceedings of the 58th Annual Meeting of the Association for Computational Linguistics}, pages 7881--7892.

\bibitem[{Shin et~al.(2020)Shin, Razeghi, Logan~IV, Wallace, and Singh}]{shin-etal-2020-autoprompt}
Taylor Shin, Yasaman Razeghi, Robert~L. Logan~IV, Eric Wallace, and Sameer Singh. 2020.
\newblock \href {https://doi.org/10.18653/v1/2020.emnlp-main.346} {{A}uto{P}rompt: {E}liciting {K}nowledge from {L}anguage {M}odels with {A}utomatically {G}enerated {P}rompts}.
\newblock In \emph{Proceedings of the 2020 Conference on Empirical Methods in Natural Language Processing (EMNLP)}, pages 4222--4235.

\bibitem[{Shinn et~al.(2023)Shinn, Cassano, Labash, Gopinath, Narasimhan, and Yao}]{shinn2023reflexion}
Noah Shinn, Federico Cassano, Beck Labash, Ashwin Gopinath, Karthik Narasimhan, and Shunyu Yao. 2023.
\newblock \href {http://arxiv.org/abs/2303.11366} {Reflexion: Language agents with verbal reinforcement learning}.

\bibitem[{Suzgun et~al.(2022{\natexlab{a}})Suzgun, Melas-Kyriazi, and Jurafsky}]{mbrd}
Mirac Suzgun, Luke Melas-Kyriazi, and Dan Jurafsky. 2022{\natexlab{a}}.
\newblock \href {https://doi.org/10.48550/ARXIV.2211.07634} {Follow the wisdom of the crowd: Effective text generation via minimum bayes risk decoding}.

\bibitem[{Suzgun et~al.(2022{\natexlab{b}})Suzgun, Melas-Kyriazi, and Jurafsky}]{suzgun-etal-2022-prompt}
Mirac Suzgun, Luke Melas-Kyriazi, and Dan Jurafsky. 2022{\natexlab{b}}.
\newblock \href {https://doi.org/10.18653/v1/2022.emnlp-main.141} {Prompt-and-rerank: A method for zero-shot and few-shot arbitrary textual style transfer with small language models}.
\newblock In \emph{Proceedings of the 2022 Conference on Empirical Methods in Natural Language Processing}, pages 2195--2222.

\bibitem[{Vernikos and Popescu-Belis(2024)}]{vernikos2024dont}
Giorgos Vernikos and Andrei Popescu-Belis. 2024.
\newblock \href {http://arxiv.org/abs/2401.06688} {Don't rank, combine! combining machine translation hypotheses using quality estimation}.

\bibitem[{Wang et~al.(2023)Wang, Wei, Schuurmans, Le, Chi, Narang, Chowdhery, and Zhou}]{wang2023selfconsistency}
Xuezhi Wang, Jason Wei, Dale Schuurmans, Quoc~V Le, Ed~H. Chi, Sharan Narang, Aakanksha Chowdhery, and Denny Zhou. 2023.
\newblock \href {https://openreview.net/forum?id=1PL1NIMMrw} {Self-consistency improves chain of thought reasoning in language models}.
\newblock In \emph{International Conference on Learning Representations}.

\bibitem[{Wei et~al.(2022)Wei, Wang, Schuurmans, Bosma, brian ichter, Xia, Chi, Le, and Zhou}]{wei2022chain}
Jason Wei, Xuezhi Wang, Dale Schuurmans, Maarten Bosma, brian ichter, Fei Xia, Ed~H. Chi, Quoc~V Le, and Denny Zhou. 2022.
\newblock \href {https://openreview.net/forum?id=_VjQlMeSB_J} {Chain of thought prompting elicits reasoning in large language models}.
\newblock In \emph{Advances in Neural Information Processing Systems}.

\bibitem[{Welleck et~al.(2023)Welleck, Lu, West, Brahman, Shen, Khashabi, and Choi}]{welleck2023generating}
Sean Welleck, Ximing Lu, Peter West, Faeze Brahman, Tianxiao Shen, Daniel Khashabi, and Yejin Choi. 2023.
\newblock \href {https://openreview.net/forum?id=hH36JeQZDaO} {Generating sequences by learning to self-correct}.
\newblock In \emph{The Eleventh International Conference on Learning Representations}.

\bibitem[{Xu et~al.(2023)Xu, Xu, Wang, Liu, Zhu, and McAuley}]{xu2023small}
Canwen Xu, Yichong Xu, Shuohang Wang, Yang Liu, Chenguang Zhu, and Julian McAuley. 2023.
\newblock \href {http://arxiv.org/abs/2305.08848} {Small models are valuable plug-ins for large language models}.

\bibitem[{Yannakoudakis et~al.(2011)Yannakoudakis, Briscoe, and Medlock}]{yannakoudakis-etal-2011-new}
Helen Yannakoudakis, Ted Briscoe, and Ben Medlock. 2011.
\newblock \href {https://aclanthology.org/P11-1019} {A new dataset and method for automatically grading {ESOL} texts}.
\newblock In \emph{Proceedings of the 49th Annual Meeting of the Association for Computational Linguistics: Human Language Technologies}, pages 180--189.

\bibitem[{Yao et~al.(2023)Yao, Zhao, Yu, Du, Shafran, Narasimhan, and Cao}]{yao2023react}
Shunyu Yao, Jeffrey Zhao, Dian Yu, Nan Du, Izhak Shafran, Karthik~R Narasimhan, and Yuan Cao. 2023.
\newblock \href {https://openreview.net/forum?id=WE_vluYUL-X} {React: Synergizing reasoning and acting in language models}.
\newblock In \emph{The Eleventh International Conference on Learning Representations}.

\bibitem[{Yasunaga et~al.(2021)Yasunaga, Leskovec, and Liang}]{yasunaga-etal-2021-lm}
Michihiro Yasunaga, Jure Leskovec, and Percy Liang. 2021.
\newblock \href {https://doi.org/10.18653/v1/2021.emnlp-main.611} {{LM}-critic: Language models for unsupervised grammatical error correction}.
\newblock In \emph{Proceedings of the 2021 Conference on Empirical Methods in Natural Language Processing}, pages 7752--7763.

\bibitem[{Zerva et~al.(2022)Zerva, Blain, Rei, Lertvittayakumjorn, C.~de Souza, Eger, Kanojia, Alves, Or{\u{a}}san, Fomicheva, Martins, and Specia}]{zerva-etal-2022-findings}
Chrysoula Zerva, Fr{\'e}d{\'e}ric Blain, Ricardo Rei, Piyawat Lertvittayakumjorn, Jos{\'e}~G. C.~de Souza, Steffen Eger, Diptesh Kanojia, Duarte Alves, Constantin Or{\u{a}}san, Marina Fomicheva, Andr{\'e} F.~T. Martins, and Lucia Specia. 2022.
\newblock \href {https://aclanthology.org/2022.wmt-1.3} {Findings of the {WMT} 2022 shared task on quality estimation}.
\newblock In \emph{Proceedings of the Seventh Conference on Machine Translation (WMT)}, pages 69--99.

\bibitem[{Zhang et~al.(2023{\natexlab{a}})Zhang, Haddow, and Birch}]{zhang2023prompting}
Biao Zhang, Barry Haddow, and Alexandra Birch. 2023{\natexlab{a}}.
\newblock Prompting large language model for machine translation: A case study.
\newblock \emph{arXiv preprint arXiv:2301.07069}.

\bibitem[{Zhang et~al.(2020)Zhang, Zhao, Saleh, and Liu}]{pegasus}
Jingqing Zhang, Yao Zhao, Mohammad Saleh, and Peter~J. Liu. 2020.
\newblock Pegasus: Pre-training with extracted gap-sentences for abstractive summarization.
\newblock In \emph{Proceedings of the 37th International Conference on Machine Learning}, ICML'20.

\bibitem[{Zhang et~al.(2023{\natexlab{b}})Zhang, Han, Zhou, Hu, Yan, Lu, Li, Gao, and Qiao}]{Zhang2023LLaMAAdapterEF}
Renrui Zhang, Jiaming Han, Aojun Zhou, Xiangfei Hu, Shilin Yan, Pan Lu, Hongsheng Li, Peng Gao, and Yu~Jiao Qiao. 2023{\natexlab{b}}.
\newblock Llama-adapter: Efficient fine-tuning of language models with zero-init attention.
\newblock \emph{ArXiv}, abs/2303.16199.

\bibitem[{Zhou et~al.(2023)Zhou, Sch{\"a}rli, Hou, Wei, Scales, Wang, Schuurmans, Cui, Bousquet, Le, and Chi}]{zhou2023leasttomost}
Denny Zhou, Nathanael Sch{\"a}rli, Le~Hou, Jason Wei, Nathan Scales, Xuezhi Wang, Dale Schuurmans, Claire Cui, Olivier Bousquet, Quoc~V Le, and Ed~H. Chi. 2023.
\newblock \href {https://openreview.net/forum?id=WZH7099tgfM} {Least-to-most prompting enables complex reasoning in large language models}.
\newblock In \emph{The Eleventh International Conference on Learning Representations}.

\end{thebibliography}

\appendix

\newpage

\section{Prompts}
We present the few-shot prompts used for all tasks.

\UseRawInputEncoding

\lstnewenvironment{customlisting}[1][]
{\lstset{
    basicstyle=\footnotesize\ttfamily, 
    frame=lines,
    framesep=1mm,
    backgroundcolor=\color{Box1Color},
    breaklines=true,
    language=TeX,
    postbreak=, 
    breakautoindent=true, 
    breakindent=0pt,
    columns=fullflexible, 
    keepspaces=true, 
    #1
}}
{}

\lstnewenvironment{customlisting2}[1][]
{\lstset{
    basicstyle=\footnotesize\ttfamily, 
    frame=lines,
    framesep=1mm,
    backgroundcolor=\color{Box2Color},
    breaklines=true,
    language=TeX,
    postbreak=, 
    breakautoindent=true, 
    breakindent=0pt,
    columns=fullflexible, 
    keepspaces=true, 
    #1
}}
{}

\lstnewenvironment{customlisting3}[1][]
{\lstset{
    basicstyle=\footnotesize\ttfamily, 
    frame=lines,
    framesep=1mm,
    backgroundcolor=\color{Box3Color},
    breaklines=true,
    language=TeX,
    postbreak=, 
    breakautoindent=true, 
    breakindent=0pt,
    columns=fullflexible, 
    keepspaces=true, 
    #1
}}
{}

\lstnewenvironment{customlisting4}[1][]
{\lstset{
    basicstyle=\footnotesize\ttfamily, 
    frame=lines,
    framesep=1mm,
    backgroundcolor=\color{Box4Color},
    breaklines=true,
    language=TeX,
    postbreak=, 
    breakautoindent=true, 
    breakindent=0pt,
    columns=fullflexible, 
    keepspaces=true, 
    #1
}}
{}

\begin{figure*}[h]
\centering
\captionsetup{justification=centering, labelfont=bf, font=small}
\begin{minipage}{\linewidth}
\begin{customlisting}
Rewrite the input sentence so that it is grammatically accurate.

Source: Yesterday as I was arriving home I saw him in your yard Mack, he looked very satisfied, Curly was laying on the grass and ... "" Yes, Oh Lord he did it because we won his team at football "Nick said."
Target: Yesterday, as I was arriving home, I saw him in your yard, Mack. He looked very satisfied, Curly was lying on the grass and ... "" Yes, oh Lord he did it because we beat his team at football ", Nick said."

Source: According to UNESCO literacy is at the heart of basic education for all and that creating literate environments andsocieties is essential for achieving the goals of eradicating poverty, reducing child mortality, curbing population growth, achieving gender equality and achieve sustainable development, peace and democracy.
Target: According to UNESCO, literacy is at the heart of basic education for all and creating literate environments and societies is essential for achieving the goals of eradicating poverty, reducing child mortality, curbing population growth, achieving gender equality and achieving sustainable development, peace and democracy.

Source: As you can see I had a very disappointing evening - the worst one during my week's holiday in London.
Target: As you can see I had a very disappointing evening - the worst one of my week's holiday in London.

Source: Anyway, to tell you the truth I'd rather take a train, for instance, it means travelling in a relaxing way, not running risks of accidents, having the chance to read or play "travelling" chess, meeting new people, as in a stage coach but moving faster
Target: Anyway, to tell you the truth, I'd rather take a train. For instance, it means travelling in a relaxing way, not running risks of accidents, having the chance to read or play "travel" chess, meeting new people, like on a stage coach but moving faster.

Source: With the advertisement, you mentioned Mr. Danny Brook and Ms. Tina Truelove were going to play but actually different people were playing, whom I have never seen before.
Target: In the advertisement, you mentioned Mr. Danny Brook and Ms. Tina Truelove were going to perform but actually different people were performing, whom I had never seen before.
\end{customlisting}
\end{minipage}
\caption{LLM prompt for GEC.}
\end{figure*}

\begin{figure*}
\captionsetup{justification=centering, labelfont=bf, font=small}
\begin{minipage}{\linewidth}
\begin{customlisting2}
Convert the set of key-value attribute pairs in the restaurant domain to a simple English-language text.


Source: name[The Eagle], eatType[coffee shop], food[English], priceRange[high], customer rating[average], area[riverside], familyFriendly[no], near[Burger King]
Target: The Eagle is near Burger King in riverside. It serves expensive English food in a coffee shop setting. It's not child friendly, but has average ratings.

Source: name[Clowns], eatType[coffee shop], food[English], customer rating[5 out of 5], area[riverside], near[Clare Hall]
Target: Clowns is a coffee shop that serves English food and is near Clare Hall. It is located riverside and has a 5 out of 5 customer rating.

Source: name[The Golden Palace], priceRange[more than £30], customer rating[high], area[city centre]
Target: The Golden Palace has a high customer rating, with meals costing more than £30. It is located in the city center.

Source: name[Wildwood], eatType[coffee shop], food[English], customer rating[1 out of 5], near[Ranch]
Target: Wildwood, English coffee shop, is situated near Ranch and has moderate pricing. It received 1 out of 5 star rating.

Source: name[Taste of Cambridge], eatType[coffee shop], food[English], area[city center], familyFriendly[yes], near[Crowne Plaza Hotel]
Target: Taste of Cambridge is a family-friendly coffee shop that serves English cuisine. It is located in the city center near Crowne Plaza Hotel.
\end{customlisting2}
\end{minipage}
\caption{LLM prompt for E2E.}
\end{figure*}

\begin{figure*}
\captionsetup{justification=centering, labelfont=bf, font=small}
\begin{minipage}{\linewidth}
\begin{customlisting3}
Write a short summary of the article in one sentence.

Source: Many more are feared trapped under rubble after hundreds of buildings collapsed. Thousands of people have been forced to take refuge in temporary shelters and mosques. Some have been left homeless after their houses were destroyed, others have fled their homes amid fears of aftershocks and a possible tsunami. Rescue workers used diggers to remove rubble in their search efforts overnight on Wednesday. Others used their bare hands and shovels to find people. A one survivors were pulled out alive on Wednesday. More than 200 buildings were either seriously damaged or toppled in the earthquake. The Pidie Jaya region, on the north Aceh coast, was the hardest hit. The tremor hit just offshore early on Wednesday morning. Many of the homes in the area have corrugated tin roofs which collapsed. Hundreds have also been rushed to the sole functioning hospital, which has been overwhelmed by patients. Banda Aceh, the provincial capital, was one of the worst hit areas by the 2004 tsunami, caused by acaused by a massive earthquake.
Target: A 6.5-magnitude earthquake struck Aceh province in Indonesia on Wednesday, killing at least 97 people.
\end{customlisting3}
\end{minipage}
\caption{LLM prompt for XSum.}
\end{figure*}

\begin{figure*}
\captionsetup{justification=centering, labelfont=bf, font=small}
\begin{minipage}{\linewidth}
\begin{customlisting4}
The disease has killed nearly 50 people and infected more than 1,400 in Tunisia. = Die Krankheit hat beinahe 50 Menschen getötet und mehr als 1400 Menschen in Tunesien infiziert.

Landray said this failure was particularly exasperating when it came to the use of convalescent plasma, which many doctors believe could have a key role to play in treating seriously ill Covid-19 patients. = Landray beklagt dieses Versagen besonders, wenn es um die Verwendung von rekonvaleszentem Plasma geht, dem laut Meinung vieler Mediziner eine wichtige Rolle bei der Behandlung ernsthaft kranker Covid-19-Patienten zukomme.

Daily cases that numbered in the hundreds dropped to low double digits. = Die täglichen Fälle, die sich auf hunderte beliefen, sanken auf zweistellige Zahlen ab.

However, a recent poll put West at two percent nationwide, neck and neck with the Libertarian Party's Jo Jorgensen and a point ahead of the Green Party's Howie Hawkins. = Jedoch lag West bei einer kürzlich erfolgten Befragung landesweit bei zwei Prozent, Kopf an Kopf mit Jo Jorgensen von der Libertarian Party und einen Punkt vor Howie Hawkins von der Green Party

Scotland's festival scene and sporting events such as the Highland games have been among those affected by restrictions brought in to prevent the spread of Covid-19 = Schottlands Festival- und Sporteventszene, wie die Highland Games, waren unter jenen die von den Einschränkungen, welche eingeführt worden sind um die Ausbreitung von Covid-19 zu verhindern, betroffen waren.
\end{customlisting4}
\end{minipage}
\caption{LLM prompt for MT.}
\end{figure*}

\end{document}